\documentclass[a4paper,twoside]{article}

\usepackage{epsfig}
\usepackage{subcaption}
\usepackage{calc}
\usepackage{amssymb}
\usepackage{amstext}
\usepackage{amsmath}
\usepackage{amsthm}
\usepackage{multicol}
\usepackage{pslatex}
\usepackage{apalike}

\usepackage{graphicx}
\graphicspath{ {plots/} }

\usepackage{SCITEPRESS}     

\begin{document}

\title{Interpreting convolutional networks trained on textual data}

\author{
\authorname{Reza Marzban\sup{1}\, Christopher John Crick\sup{1}}
\affiliation{\sup{1}Computer Science Department,  Oklahoma State University,  Stillwater, OK, USA}
\email{reza.marzban@okstate.edu, chriscrick@cs.okstate.edu}
}


\abstract{There have been many advances in the artificial intelligence field due
to the emergence of deep learning. In almost all
sub-fields, artificial neural networks have reached or exceeded
human-level performance. However, most of the models are not
interpretable. As a result, it is hard to trust their decisions,
especially in life and death scenarios. In recent years, there has
been a movement toward creating explainable artificial intelligence,
but most work to date has concentrated on image processing models, as
it is easier for humans to perceive visual patterns. There has been
little work in other fields like natural language processing. In
this paper, we train a convolutional model on textual data and analyze
the global logic of the model by studying its filter values. In the
end, we find the most important words in our corpus to our model’s
logic and remove the rest (95\%). New models trained on just the 5\%
most important words can achieve the same performance as the original
model while reducing training time by more than half. Approaches such
as this will help us to understand NLP models, explain their decisions
according to their word choices, and improve them by finding
blind spots and biases.
}

\onecolumn \maketitle \normalsize \setcounter{footnote}{0} \vfill

\section{\uppercase{Introduction}}
\label{sec:introduction}

\noindent 
In the big data era, traditional naive statistical models and machine
learning algorithms are not able to keep up with the growth in data
complexity.  Such algorithms are the best choice when our data size is
limited and nicely shaped in tabular formats. Previously, we were
interested in analyzing structured data in databases, inserted by an
expert, but now we use machine learning in every aspect of life.  Most
of the data is unstructured, such as images, text, voice, and
videos. In addition, the amount of data has increased
significantly. Traditional machine learning algorithms cannot handle
these types of data at our desired performance level.  Artificial
Neural Networks (ANNs) have undergone several waves of popularity and
disillusionment stretching back to 1943.  Recently, due to increases
in computing power and data availability, the success and improvement
of ANNs and deep learning have been the hot topic in machine learning
conferences.  In many problem areas, deep learning has reached or
surpassed human performance, and is the current champion in fields
from image processing and object detection to Natural Language
Processing (NLP) and voice recognition.

Although deep learning has increased performance across the board, it
still has many challenges and limitations. One main criticism of deep
learning models is that they are black boxes: we throw data at it, use
the output and hope for the best, but we do not understand how or why.
This is less likely to be the case with traditional statistical and
machine learning methods such as decision trees.  Such algorithms are
interpretable and easy to understand, and we can know the reasoning
behind the decisions they make. Explainability is very important if we
want people to rely on our models and trust their decisions. This
becomes crucial when the models' decisions are life-and-death
situations like medicine or autonomous vehicles, and it is the reason
behind the trend toward explainable artificial intelligence.  In
addition to boosting users' trust, however, interpretability also
helps developers, experts and scientists learn the shortcomings of
their models, check them for bias and tune them for further
improvement.

Many papers and tools have recently contributed to explainability in
deep learning models, but most of them have concentrated on machine
vision problems, as images are easier to visualize in a 2-dimensional
space. They can use segmentation and create heatmaps to show users
which pixel or object in the image is important, and which of them
caused a specific decision.  Humans can easily find visual patterns in
a 2-dimensional space.  However, there is also a need for model
interpretability in other contexts like NLP, the science of teaching
machines to communicate with us in human-understandable languages. As
NLP data mostly consists of texts, sentences and words, it is very
hard to visualize in a 2-D or 3-D space that humans can easily
interpret, even though visualization of a model is an important part
of explainable AI.

Deep learning networks have various architectures. Convolutional
neural networks (CNNs) were primarily designed for image
classification but can be applied to all types of data. Recurrent
neural networks (RNNs) are often a good choice for time-series data;
Long Short Term Memory (LSTMs) and Gated Recurrent Units (GRUs) are
common forms of RNN.  Many scientists prefer to use LSTMs on NLP
problems, as they are constructed with time series in mind. Each word
can be looked at as a time step in a sentence. 1-dimensional CNNs can
also be used on textual data. They are faster than LSTMs and perform
well on well-known NLP problems like text classification and sentiment
analysis.

In this paper we have created a simple 1-D CNN and trained it on a
large labeled corpus for sentiment analysis, which aims understand the
emotion and semantic content of text and predict if its valence is
positive or negative. We then deconstructed and analyzed the CNN layer
filters. We tried to understand the filter patterns and why the
learning algorithm would produce them.  The first result indicates
that the filter weights cover around 70\% of a layer's information,
while their order covers only 30\%. As a result, randomly shuffling
filters causes a particular layer to lose 30\% of the accuracy
contributed by a particular layer.  We also used activation
maximization to create an equation to find the importance of each word
in our corpus dictionary to the whole model. This importance rate is
not specific to a single decision but to the whole model logic.  We
were able to order the word corpus according to their decision-making
utility, and used this information to train a new model from scratch
on only the most important words. We observed that a model built from
the most important 5\% of words is just as accurate as one that uses
the whole corpus, but input size and training time are
reduced significantly.

\section{\uppercase{Related Work}}
\label{sec:related_work}
\noindent 
Explainable artificial intelligence (XAI)
\cite{gunning2017explainable} helps us to trust AI models, improve
them by identifying blind spots or removing bias. It involves creating
explanations of models to satisfy non-technical users
\cite{du2019techniques}, and helps developers to justify and improve
their models. Such models come in various
flavors \cite{adadi2018peeking}; they can provide local explanations
of each prediction or globally explain the model as a whole.
Layer-wise relevance propagation (LRP)
\cite{bach2015pixel,montavon2018methods} matches each prediction in
the model to the input features that have caused it. LIME
\cite{ribeiro2016should} is a method for providing local
interpretable model-agnostic explanations.  These techniques help us
to trust deep learning models.

Most of the XAI community has concentrated on image processing and
machine vision as humans find it easy to understand and find patterns
in visual data.  Such research has led to heat maps, saliency maps
\cite{simonyan2013deep} and attention networks
\cite{wang2017residual}.  However, other machine learning fields,
such as NLP, have not yet seen nearly as many research efforts.  There
have been many improvements in NLP models' performance in recent years
\cite{collobert2011natural} but very few of them concentrate on
creating self-explanatory models.

Arras \cite{arras2017relevant} tried to find and highlight the words in a
sentence leading to a specific classification using LRP and identified
the words that vote out the final prediction.  This can help identify
when a model arrives at a correct prediction through incorrect logic
or bias, and provide clues toward fixing such errors.  These kinds of
local explanations help to confirm single model predictions, but
methods for understanding the global logic of models and specific
architectures are necessary to provide insight for improving future
models. Such techniques are model-specific and dependent on the
architecture used.

It is a common belief that RNNs, and specifically LSTMs
\cite{hochreiter1997long} are efficient for NLP tasks. 1-dimensional
CNNs are also used for common NLP tasks like sentence classification
\cite{kim2014convolutional} and modeling
\cite{kalchbrenner2014convolutional}. Le \cite{le2018convolutional}
shows how CNN depth affects performance in sentiment analysis.
Yin \cite{yin2017comparative} compares the performance of RNNs and CNNs
on various NLP tasks. Wood \cite{wood2018convolutions} shows that CNNs
can outperform RNNs on textual data, in addition to being faster.

There have been many works trying to interpret and visualize CNN
models. Most of them tried to visualize the CNNs on famous visual
object recognition databases like
ImageNet \cite{zeiler2014visualizing}. Four main methods have been
used to visualize models in image processing tasks: activation
maximization, network inversion, deconvolutional neural networks, and
network dissection
\cite{qin2018convolutional}. Yosinski \cite{yosinski2015understanding} has
created tools to visualize features at each layer of a CNN model in
image space.  Visualization and interpretation for other types of
data, such as text, are nowhere near as well-developed, but there have
been a few attempts. Choi \cite{choi2016explaining} tried to explain a CNN
model that classifies genres of music, and showed that deeper layers
capture textures. Xu \cite{xu2015show} used attention-based models to
describe the contents of images in natural language, showing saliency
relationships between image contents and word generation.

The most challenging part in visualizing NLP models is that after
tokenizing textual data with available tools like
NLTK \cite{bird2009natural}, each token or word is represented by an
embedding \cite{maas2011learning}, \cite{mikolov2013distributed},
 \cite{rehurek2010software}. An embedding is a vector of numbers that
represent a word's semantic relationship to other words.  Pre-trained
embeddings like GloVe \cite{pennington2014glove} are available that
are trained on a huge corpus. However, they are not understandable by
humans, and it is very hard to explain models that are built upon
them. Li \cite{li2015visualizing} introduced methods illustrate the
saliency of word embeddings and their contribution to the overall
model's comprehension. Rajwadi \cite{rajwadi2019explaining} created a 1-D
CNN for a sentiment analysis task and used a deconvolution technique
to explain text classification. They estimate the importance of each
word to the overall decision by masking it and checking its effect on
the final classification score.

In our paper, we also create a 1-D CNN for sentiment analysis on the
IMDb dataset \cite{maas2011learning}. However, instead of creating a
local explanation for each prediction and decision, we describe the
whole model's logic and try to explain it in a layer-wise manner by
studying the filters of the trained model.

\section{\uppercase{Technical Description}}
\label{sec:tech_desc}

\subsection{Dataset Introduction and Preprocessing}
\label{Preprocessing}
\noindent 
The dataset used throughout this paper is the IMDb Large Movie Review
Dataset \cite{maas2011learning}, which is a famous benchmark for NLP
and sentiment analysis in particular. It contains around 50,000
balanced, labeled reviews, rated either positive or negative (no
neutral reviews are included). We split the data into training and
validation sets with a ratio of 90:10 respectively.

In our preprocessing step, we used NLTK \cite{bird2009natural} to
tokenize the reviews, remove punctuation, numerical values, HTML tags
and stop-words.  We also removed words that are not in the English
dictionary (like typos and names).  We set the sequence length to 250:
sentences longer than that were truncated, while shorter ones were
padded with zeros.  After preprocessing, the corpus dictionary
contained 23,363 words.

To transform the textual data into numerical values that can be used
by our models, we used Word2Vec \cite{rehurek2010software} to create
100-dimensional embeddings for each word.  This high-dimensional space
is intended to represent semantic relationships between each word.
Each review is therefore represented as a $250 \times 100$ matrix and
a binary target value.

\subsection{Basic CNN Model Setup}
\label{basic_cnn}
\noindent 
We use a 1-dimensional CNN for the sentiment analysis problem. The
architecture is presented in Table \ref{tab:CNN}.  The embedding layer
contains parameters for each of 100 dimensions for each word in the
corpus, plus one (for unknown words).  The embeddings are untrainable
in the CNN, having been trained in an unsupervised manner on our
corpus, and we did not want to add an extra variable to our
evaluation.  The first convolutional layer has 32 filters with size of
5 and stride of 1. The second has 16 filters with size of 5 and stride
of 1. Both max-pooling layers have size and stride of 2.  All
computational layers use the ReLu activation function.  The output
layer's activation function is sigmoid (as our target is binary).  Our
models are trained for 5 epochs with a decaying learning rate of
0.001.

\begin{table}
\caption{Basic CNN Model.}
\label{tab:CNN}
\begin{center}
\begin{small}
\begin{sc}
\renewcommand{\arraystretch}{1.15}
\begin{tabular}{lcr}
\multicolumn{1}{c}{\textbf{\begin{tabular}[c]{@{}c@{}}Layer\\ Type\end{tabular}}} & \textbf{\begin{tabular}[c]{@{}c@{}}Output\\ Shape\end{tabular}}& \multicolumn{1}{c}{\textbf{\begin{tabular}[c]{@{}c@{}}Number of\\ Parameters\end{tabular}}} \\ \hline
Input & (?,250) & 0 \\ \hline
Embedding & (?,250,100) & 2,336,400 \\ \hline
Convolutional-1 & (?,246,32) & 16,032 \\ \hline
Max Pooling-1 & (?,123,32) & 0 \\ \hline
Convolutional-2 & (?,119,16) & 2,576 \\ \hline
Max Pooling-2 & (?,59,16) & 0 \\ \hline
Flatten & (?,944) & 0 \\ \hline
Fully connected & (?,128) & 120,960 \\ \hline
Output & (?,1) & 129 \\ \hline \hline
\multicolumn{2}{c}{\textbf{Total Parameters}} &2,476,097 \\ \hline
\multicolumn{2}{c}{\textbf{Trainable Parameters}} &139,697 \\ \hline
\multicolumn{2}{c}{\textbf{Non-Trainable Parameters}} &2,336,400 \\ \hline
\end{tabular}
\end{sc}
\end{small}
\end{center}
\end{table}

\subsection{Analyzing and Interpreting Convolutional Layer Filters}
\label{analyse_cnn_filters}
\noindent 
In order to understand the logic of our CNN model, we studied the
first convolutional layer's filter weights.  We created three new
models with identical architecture to our baseline model. In two of
these new models, we copy the weights of the basic model's first
convolutional layer and make them untrainable, then initialize the
rest of the layers randomly and train normally.  We then shuffled the
filter weights in the first layer, either within each filter or across
the whole set of filters.  In the last model, we randomly initiate the
first layer's weight and freeze it.



\subsection{Word Importance through Activation Maximization}
\label{word-importance}
\noindent 
It is difficult to analyze the actual values learned by the
convolutional filters.  If we cannot interpret them as they are, we
are not able to follow the reasoning behind a model's decisions,
either to trust or to improve them.  As a result, we wanted to
concentrate on the input space, and check each word's importance to
our model. Previous research has focused on finding significant words
that contribute to a specific decision.  This is helpful, but it only
demonstrates local reasoning specific to a single input and the
model's decision in that instance. We are interested in global
explanations of the model, so that the model can convey its overall
logic to users. To do so on our CNN model, we applied Equation
\ref{equation1}, which provides an importance rating for each word,
according to our first convolutional layer's filters.

\begin{equation}\label{equation1}
\begin{split}
& \text{importance} = \\ & \left \{ \sum_{f=1}^{F}\sum_{s=1}^{S}\sum_{i=1}^{I} \left | w_{i}*\text{Filter}_{f*s*i} \right | | w\in \text{Corpus, Filter} \right \}
\end{split}
\end{equation}

In equation \ref{equation1}, $F$ is the number of filters, $S$ is the
size of filters, and $I$ is the embedding length. $w$ is a word
embedding vector with a length of $I$.  Corpus is a matrix of our
entire word embedding of size $m*I$, in which $m$ is the count of
unique words in our corpus dictionary.  Filter is a 3-D tensor of size
$F*S*I$. This equation calculates the sum of activations of all
filters caused by a single word.  In our models, Corpus contains
$13,363 \times 100$ elements, each $w$ is a vector of 100 numbers, and
the Filter size is $32 \times 5 \times 100$.

The above equation can be used to compute an importance rating for
each and every word in our corpus according to our model's logic. One
of the benefits of studying these ratings is that we can understand
what types of words affect our model's decisions most strongly.  To
investigate this further, we dropped unimportant words and trained new
models on a subset of data containing just the most important
vocabulary. By doing so, we learn from our basic model and can inject
its insights to new models, to develop faster and better-performing
ones.  In order to prove our hypothesis, we created a new model
trained on 5\% of the most important words and compared its
performance and training time to three baseline models. In all of
these models the architecture is the same and we train the embedding
weights as well. Our first baseline model uses 100\% of the words in
the corpus, our second uses 5\% of words chosen randomly, and the
third uses all of the words \emph{except} the 5\% most important, in
other words, the least important 95\% of words.

\section{\uppercase{Experimental Results}}
\label{sec:results}

\subsection{Performance of Models with Shuffled Filters}
\label{shuffled}

\begin{table*}
\caption{Comparison of models with shuffled filters.  Accuracy improvement represents the increase in test accuracy compared to the following model.}
\label{tab:shuffled}
\begin{center}
\begin{sc}
\renewcommand{\arraystretch}{1.25}
\begin{tabular}{lccc}
\textbf{}                                  & \textbf{Train Accuracy} & \textbf{Test Accuracy} & \textbf{Accuracy improvement} \\ \hline
Basic Model                       & 93.24                   & 83.19                  & 1.82                                  \\ \hline
Shuffle within filters            & 90.18                   & 81.37                  & 2.92                                  \\ \hline
Shuffle across filters            & 87.64                   & 78.45                  & 10.52                                 \\ \hline
First layer random initialization & 81.11                   & 67.93                  & 17.93                                 \\ \hline
\end{tabular}
\end{sc}
\end{center}
\end{table*}

\noindent 
After creating three models with the same architecture as our basic
model, we set their first convolutional layer weights and make them
untrainable. The rest of the model is trained normally for five
epochs.  Table \ref{tab:shuffled} shows that, when the first layer is
assigned randomly and then frozen, accuracy is around 68\%, 18\%
higher than random prediction (as our target variable is binary and
balanced).  Even when the first layer does not learn anything or
contribute to the classification outcomes, the rest of the model
learns enough for modest success.

When we train the first layer normally (in the basic model), it
contributes around 15\% to overall performance. 2/3 of this
contribution belongs to the filter value choices and 1/3 belongs to
the order of the sequence in our filters. That is the reason that when
we shuffle all 160 (32 filters $\times$ 5 units in each filter)
weights across all filters, only around 5\% of overall accuracy is
lost.

Based on these experimental results, we learned that the ordering of
each filter is much less important, compared to the crucial filter
values found by a model. In addition, we also learned that the relationship
between neighboring filter values is not especially strong, since not
much performance is lost if the positions of each value are randomized
throughout the convolutional layer.

\begin{figure*}
  \centering
  \centerline{\includegraphics[width=\linewidth]{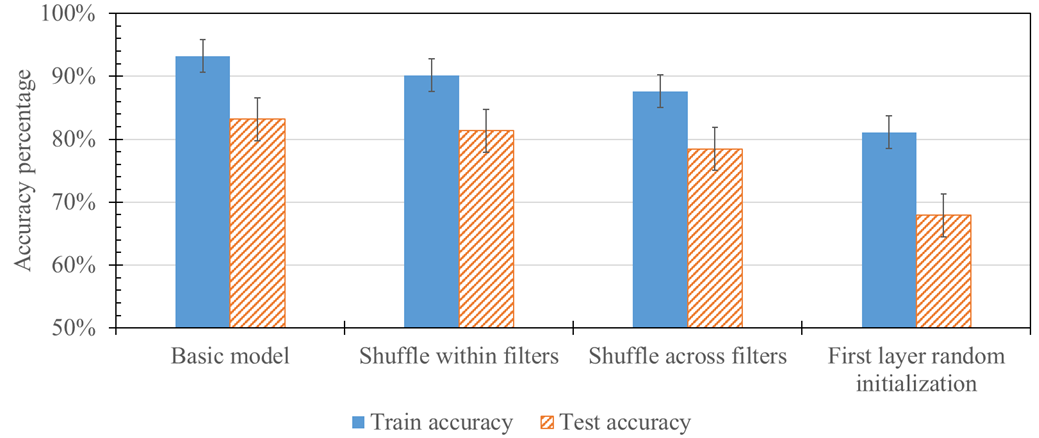}}
\caption{Comparison of models with shuffled filters.}
\label{fig:shuffled}
\end{figure*}

\subsection{Clustering on Words and Filters}
\label{clustering}

\begin{table*}
\caption{Clustering results.}
\label{tab:clustering}
\begin{center}
\begin{small}
\begin{sc}
\renewcommand{\arraystretch}{1.25}
\begin{tabular}{crrr}
\textbf{K}    & \multicolumn{1}{c}{\textbf{\begin{tabular}[c]{@{}c@{}}Sum of squared\\ distances\end{tabular}}} & \multicolumn{1}{c}{\textbf{\begin{tabular}[c]{@{}c@{}}Count of elements in\\ most populated cluster\end{tabular}}} & \multicolumn{1}{c}{\textbf{\begin{tabular}[c]{@{}c@{}}Percent of elements in\\ most populated cluster\end{tabular}}} \\ \hline
1    & -                                                                                                & 23363                                                                                                               & 100.00                                                                                                                \\ \hline
5    & 218924                                                                                           & 19869                                                                                                               & 85.04                                                                                                                 \\ \hline
10   & 204173                                                                                           & 14507                                                                                                               & 62.09                                                                                                                 \\ \hline
20   & 191238                                                                                           & 11871                                                                                                               & 50.81                                                                                                                 \\ \hline
100  & 155071                                                                                           & 8433                                                                                                                & 36.09                                                                                                                 \\ \hline
200  & 134379                                                                                           & 7472                                                                                                                & 31.98                                                                                                                 \\ \hline
500  & 98158                                                                                            & 5210                                                                                                                & 22.30                                                                                                                 \\ \hline
1000 & 63640                                                                                            & 4936                                                                                                                & 21.13                                                                                                                 \\ \hline
2000 & 32948                                                                                            & 2486                                                                                                                & 10.64                                                                                                                 \\ \hline
\end{tabular}
\end{sc}
\end{small}
\end{center}
\end{table*}

\begin{figure}
  \centering
  \centerline{\includegraphics[width=1.1\linewidth]{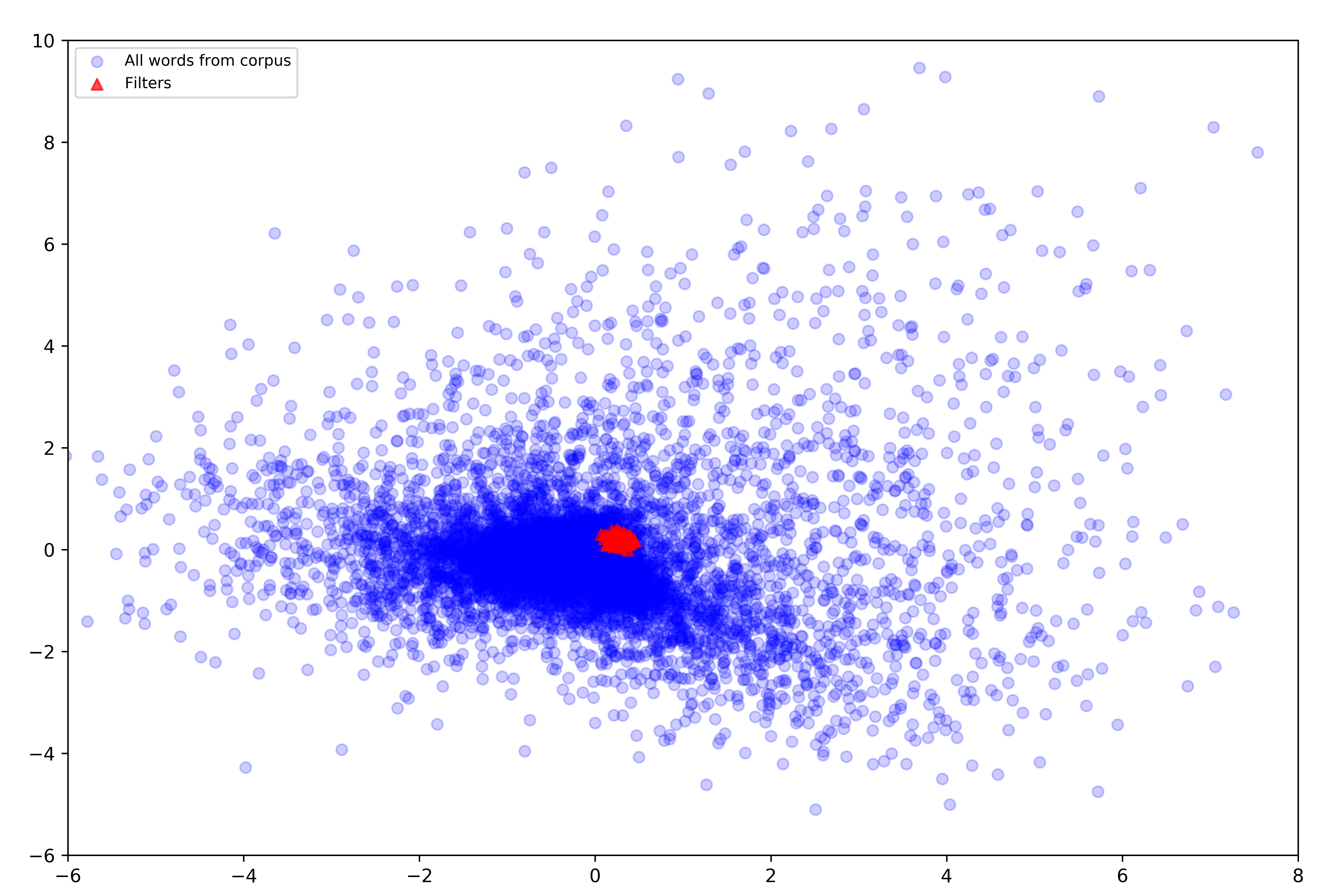}}
\caption{Words vs filters}
\label{fig:words_vs_filters}
\end{figure}

\noindent 
Clustering is an efficient way to understand patterns within data. To
investigate such patterns, we concatenated all of our word embeddings
(23,363 $\times$ 100) and our filters (160 $\times$ 100) to create $k$
clusters. We tested different $k$ between 2 to 2000. The result of the
clustering can be seen in Table \ref{tab:clustering}.  No matter which
size $k$ we choose, there is a single crowded cluster that contains
most of the words (e.g., when we produce five clusters, 85 percent of
words belong to one cluster). The most crowded cluster contains all
160 filter values.  This means that in word embedding space, most of
the words are concentrated in a small part of space, and our model
chose our filters to be in that space as well.  Figure
\ref{fig:words_vs_filters} shows how tightly the filter values are
concentrated in the main cluster of words (using PCA to represent
100-dimensional embedding vectors in two dimensions).

\subsection{Performance of models on most important words}
\label{important_words}
\noindent 
After finding the importance rating of every single word in our corpus
according to Equation \ref{equation1} , we created six new models.
All of them share the same architecture as our basic model shown in
Table \ref{tab:CNN}, and each of them is trained only on a subset of
the corpus of words.  We choose the top $n$ most important words in
our corpus and drop the rest. We train our brand new models on these
subsets of words from scratch (weights randomly initiated).  Even
after dropping 95\% of words and training a new model just on 5\% of
the most important, the performance does not decrease
significantly. The performance of these models is presented in Table \ref{tab:top_n}.

\begin{table*}
\begin{center}
\caption{New models trained on subset of words.}
\label{tab:top_n}
\begin{small}
\begin{sc}
\renewcommand{\arraystretch}{1.25}
\begin{tabular}{cccc}
\textbf{Words kept percentage} & \textbf{Word counts} & \textbf{Train accuracy} & \textbf{Test accuracy} \\ \hline
100.0 (Base Model)                                   & 23,363                                    & 93.24                                        & 83.19                                       \\ \hline
80.0                                                 & 18,691                                    & 92.71                                        & 83.16                                       \\ \hline
50.0                                                 & 11,682                                    & 93.43                                        & 83.14                                       \\ \hline
10.0                                                 & 2,337                                     & 92.83                                        & 83.67                                       \\ \hline
5.0                                                  & 1,169                                     & 92.34                                        & 82.53                                       \\ \hline
1.0                                                  & 234                                       & 87.02                                        & 78.16                                       \\ \hline
0.5                                                  & 117                                       & 84.75                                        & 74.62                                       \\ \hline
\end{tabular}
\end{sc}
\end{small}
\end{center}
\end{table*}

\begin{figure}
    \centering
    \includegraphics[width=1.1\linewidth]{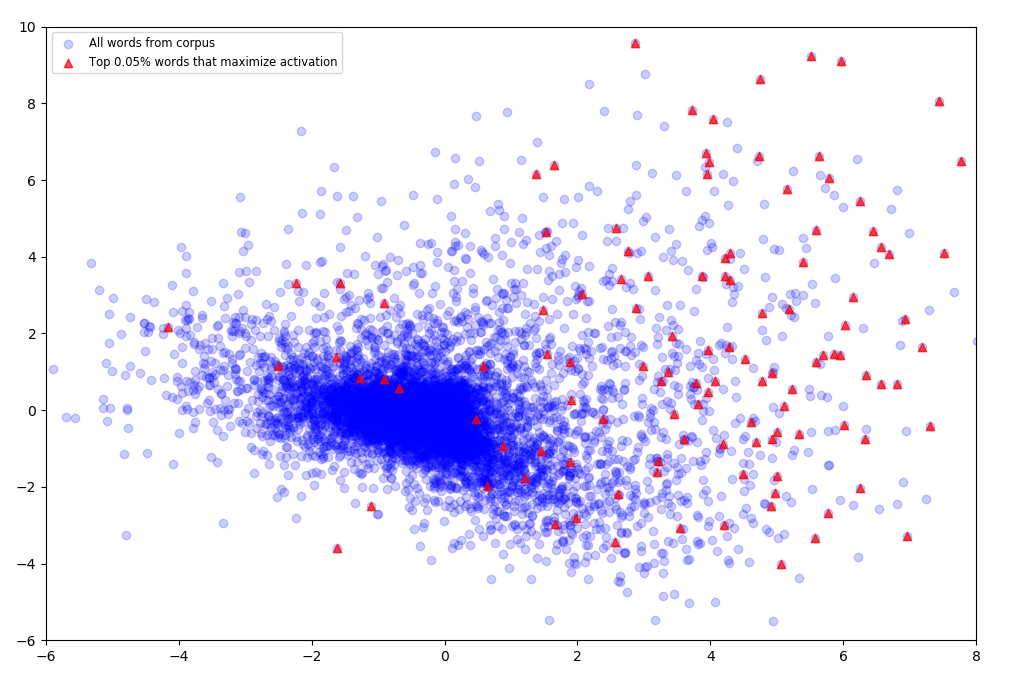}
  \caption{Most important words vs all words.}
  \label{fig:top_words1}
\end{figure}

\begin{figure}
    \centering
    \includegraphics[width=1.1\linewidth]{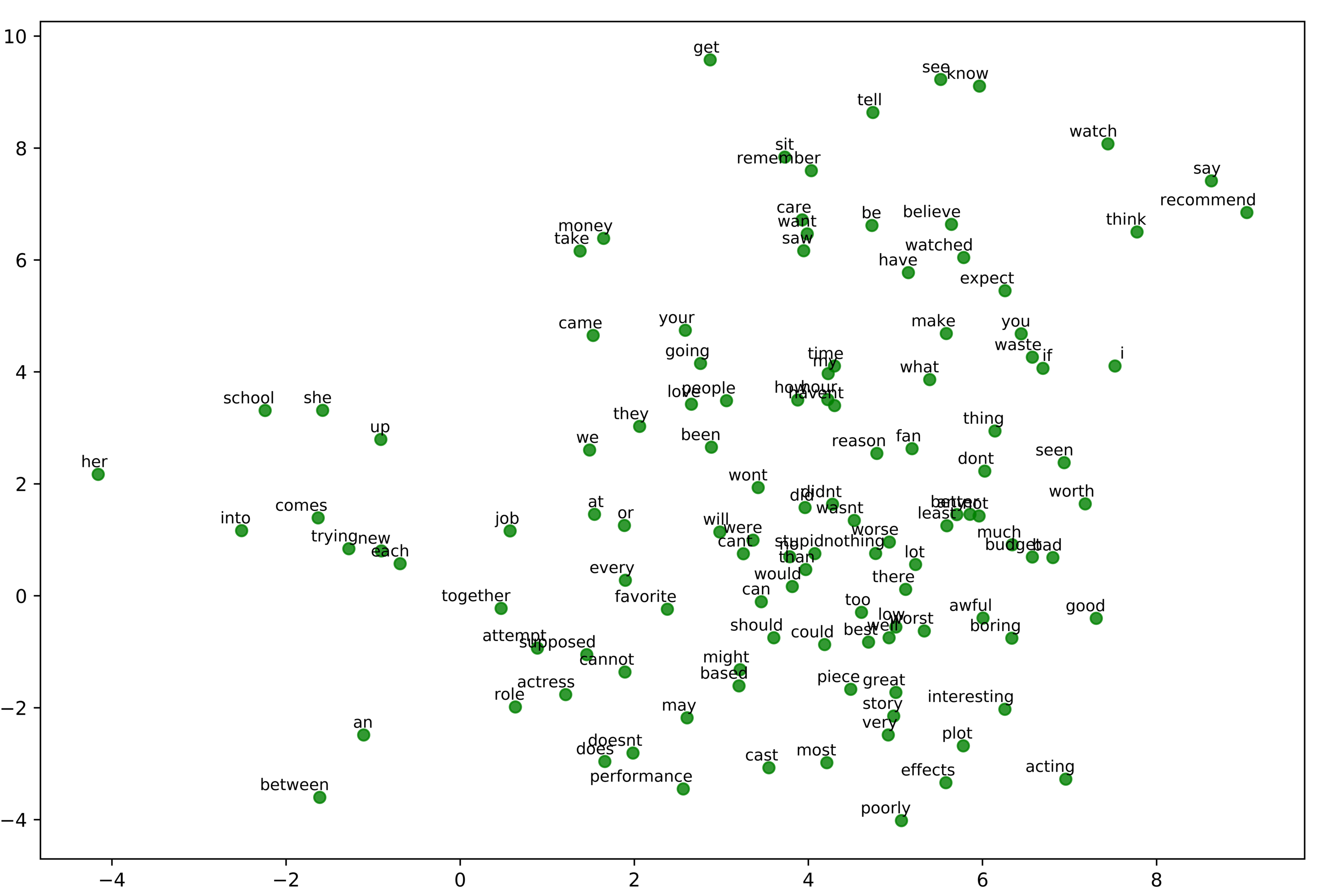}
  \caption{Most important words.}
  \label{fig:top_words2}
\end{figure}

\begin{figure*}
  \centering
  \centerline{\includegraphics[width=\linewidth]{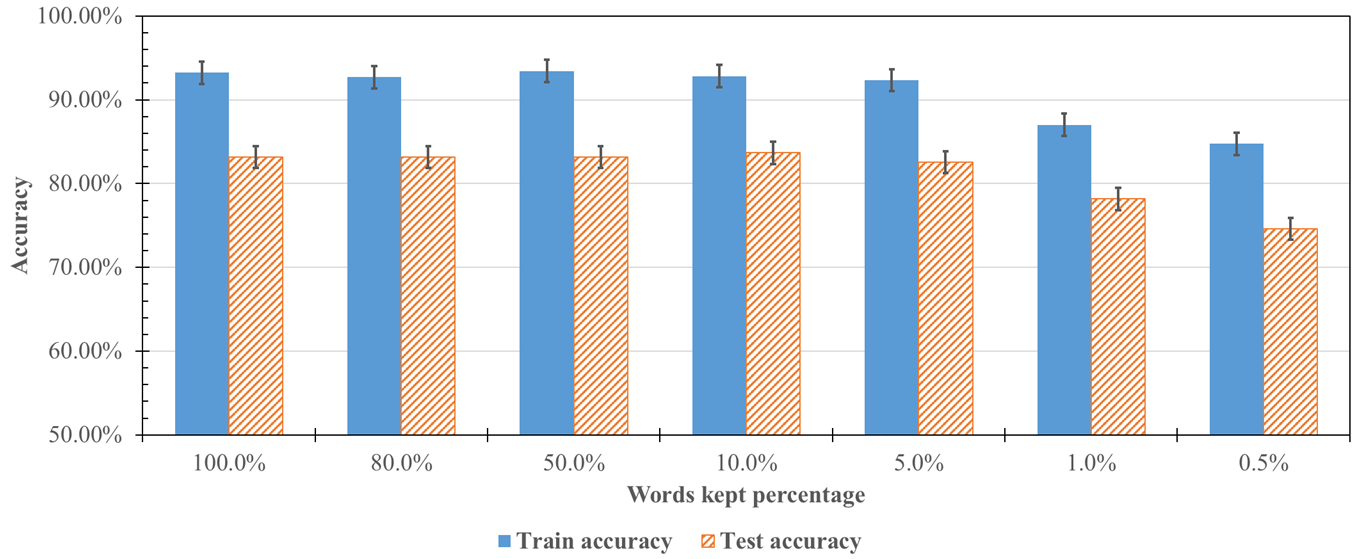}}
\caption{Comparison of models trained on most important words}
\label{fig:model_comparison}
\end{figure*}

Although the model does start to lose information, and thus
classification performance, once the corpus is reduced to 1\% of its
original size or below, performance remains strong when using only 5\%
of available words.  Our final set of experiments focus on this
behavior.  In figures \ref{fig:top_words1} and \ref{fig:top_words2},
we used PCA to reduce the dimensionality of the word embeddings from
100 to 2 in order to represent words in 2-D space, and show how the
most important words form a reasonable coverage of the space. 
Figure \ref{fig:top_words1} shows the important words as a fraction of all of the words, and
figure \ref{fig:top_words2} shows which words were found to represent the embedding
space.  The figures are separated for clarity.  To compare, we
established three baseline models: one which uses all words in our
corpus, one that also uses 5\%, but selected randomly rather than via
Equation \ref{equation1}, and one that uses all \emph{except} the most
important 5\% of words.  Results are shown in Table
\ref{tab:baseline}.  Our selected words perform much better than
randomly choosing the 5\% of words (a 20\% increase in test accuracy).

\par
\begin{figure*}
  \centering
  \centerline{\includegraphics[width=400pt]{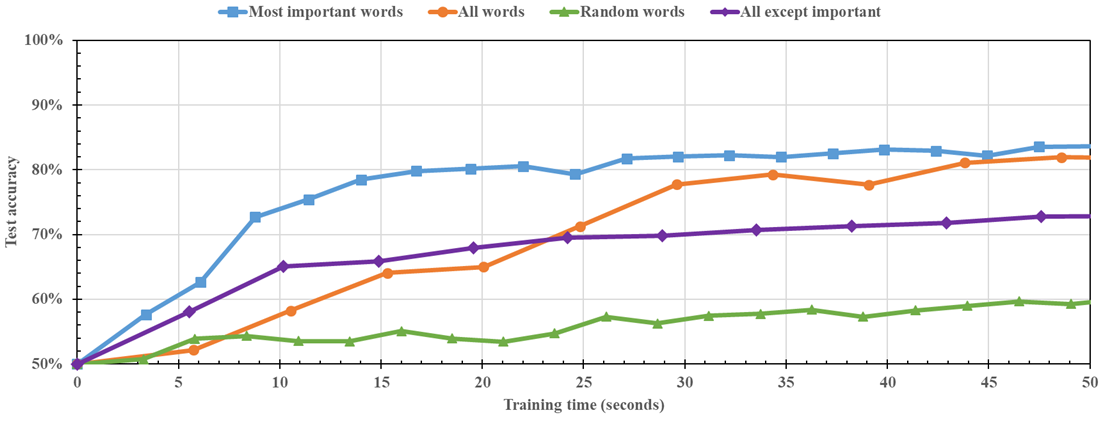}}
\caption{Comparing our model with three baseline models based on testing accuracy and training time - points in each line represent 1/20th of an epoch}
\label{fig:baseline}
\end{figure*}

 Table \ref{tab:baseline} and Figure \ref{fig:baseline} also show
that restricting the model to the most important words results in much
faster performance than the model using every available word in the
corpus.  Whether measured in epochs or seconds, the restricted model
is more than twice as fast at learning.  Unsurprisingly, speed is
equivalent between both models which use only 5\% of the data, but one
that uses the important words performs much better.  If the best 5\%
of words identified via Equation \ref{equation1} are eliminated, the
model has the worst of both worlds and is neither fast nor accurate.

\begin{table*}
\caption{Comparing new models to baseline models}
\label{tab:baseline}
\begin{center}
\begin{small}
\begin{sc}
\renewcommand{\arraystretch}{1.25}
\begin{tabular}{lcccccr}
\multicolumn{1}{c}{\textbf{Model name}} & \textbf{\begin{tabular}[c]{@{}c@{}}\% words \\ used\end{tabular}} & \textbf{\begin{tabular}[c]{@{}c@{}}Word\\ count\end{tabular}} & \textbf{\begin{tabular}[c]{@{}c@{}}Train \\ accuracy\end{tabular}} & \textbf{\begin{tabular}[c]{@{}c@{}}Test\\ accuracy\end{tabular}} & \textbf{\begin{tabular}[c]{@{}c@{}}Average epoch \\ training time\\ (seconds)\end{tabular}} & \multicolumn{1}{c}{\textbf{\begin{tabular}[c]{@{}c@{}}Number of\\ parameters\end{tabular}}} \\ \hline
Most important words                       & 5\%                                                              & 1,169                                                         & 93.67\%                                                            & 84.42\%                                                          & 36.3062                                                                                   & 256,697                                                                                      \\ \hline
All words                    & 100\%                                                            & 23,363                                                        & 96.87\%                                                            & 85.33\%                                                          & 81.6395                                                                                   & 2,476,097                                                                                    \\ \hline
Random words                      & 5\%                                                              & 1,169                                                         & 70.68\%                                                            & 64.29\%                                                          & 36.1543                                                                                   & 256,697                                                                                      \\ \hline
All except important                       & 95\%                                                              & 22,194                                                         & 83.86\%                                                            & 74.52\%                                                          & 79.6917 & 2,359,197                                                                                      \\ \hline
\end{tabular}
\end{sc}
\end{small}
\end{center}
\end{table*}

\section{\uppercase{Conclusion And Future Works}}
\label{conclusion}
\noindent 
The field of machine learning has long focused on how to improve the
performance of our models, identify useful cost functions to optimize,
and thereby increase prediction accuracy.  However, now that
human-level performance has been reached or exceeded in many domains
using deep learning models, we must investigate other important
aspects of our models, such as explainability and interpretability.
We would like to be able to trust artificial intelligence and rely on
it even in critical situations.  Furthermore, beyond increasing our
trust in a model's decision-making, a model's interpretability helps us
to understand its reasoning, and it can help us to find its weaknesses
and strengths.  We can learn from the model’s strengths and inject
them into new models, and we can overcome their weak points by
removing their bias.

In this paper, we investigated the logic behind the decisions of a 1-D
CNN model by studying and analyzing its filter values and determining
the relative importance of the unique words within a corpus
dictionary.  We were able to use the insights from this investigation
to identify a small subset of important words, improving the learning
performance of the training process by better than double.  Future
work includes expanding these techniques to investigate structures
beyond the first layer of the convolutional network.  In addition, we
are planning to deepen our study of the ability of our model to
identify important words.  By performing sensitivity analysis,
alternately verifying or denying the model's access to words it deems
vital, we will hopefully be able to facilitate the transfer of
linguistic insights between human experts and learning systems.

\bibliographystyle{apalike}
{\small
\bibliography{ICPRAM_2021_bib}}

\end{document}